# Extended Intelligence

**David Barack**  dbarack@gmail.com
*University of Pennsylvania*

**Andrew Jaegle**  drewjaegle@deepmind.com
*DeepMind*

**Abstract**

We argue that intelligence — construed as the disposition to perform tasks successfully—is a property of systems composed of agents and their contexts. This is the thesis of extended intelligence. We argue that the performance of an agent will generally not be preserved if its context is allowed to vary. Hence, this disposition is not possessed by an agent alone, but is rather possessed by the system consisting of an agent and its context, which we dub an agent-in-context. An agent's context may include an environment, other agents, cultural artifacts (like language, technology), or all of these, as is typically the case for humans and artificial intelligence systems, as well as many non-human animals. In virtue of the thesis of extended intelligence, we contend that intelligence is context-bound, task-particular and incommensurable among agents. Our thesis carries strong implications for how intelligence is analyzed in the context of both psychology and artificial intelligence.

**Keywords:** Artificial Intelligence, Psychology, Biological Intelligence, Philosophy of Science, Dispositions

I. **Introduction**

Intelligence is a disposition to perform tasks. Because the grounds of this disposition extend beyond the agent, we argue that intelligence is properly a property of agents-in-contexts. We call this the thesis of extended intelligence:

(EI): Intelligence is a property of agents-in-contexts.

In this paper, we argue for (EI). Specifically, we will argue that intelligence is a property of a system that goes beyond the agent to include some aspects of the agent's context. We also discuss three implications of the thesis: intelligence is context-bound, task-particular and incommensurable.

The standard analysis of intelligence in AI [1, 2, 3, 4] and psychology [5, 6] assumes that individual agents are intelligent, can perform many different tasks, and can be compared (in counterpoint cf. [7], and for a historical review of the use of this term in psychology and psychometrics, see [8], p.66-84). In this view, the goal of AI research is to engineer systems that are intelligent in virtue of possessing some general capacity to perform tasks (and which are comparable to humans to the extent that they possess it). AI analyses of intelligence typically appeal to the design or analysis of individual systems for performing any task [9], that can consistently produce behavior indistinguishable from that of humans in similar settings [1, 4], or that can learn or iteratively improve on timescales as fast or faster than a typical or expert human individual [3, 10]. The standard analysis from human psychology is similar insofar as it assesses the human capacity to perform many different tasks [5, 6]. The goal of the psychological study of human intelligence is to assess the degree to which humans possess this single general capacity and to compare humans to any other agent.

These analyses assume that intelligence is (i) individualistic: intelligence is a property of individual agents, (ii) general: intelligence refers to or measures an individual's capacity to perform tasks in any context, and (iii) commensurable: all intelligences can be measured using a single scale. These assumptions commonly drive researchers to draw the following conclusions: the goal of AI research is to design and engineer systems that are intelligent in virtue of possessing some single, general capacity to perform tasks (and which are comparable to humans to the extent that they possess it); the goal of the psychological study of human intelligence is to analyze and assess the degree to which humans possess this single general capacity and to compare specific humans to conspecifics or other agents.

In contrast, the sciences of animal behavior, the brain, and cognition, and certain research programs in artificial life and AI [11], involve descriptions of systems that have developed capacities and behaviors to adapt to opportunities and adversities imposed by ecological and sociocultural constraints. Empirical observations suggest that intelligence is (i) sociocultural: found in particular sociocultural contexts, for example in societies composed of individuals with a wide range of behaviors and competences [12]; (ii) particular:

variable in the level of generality and level of sophistication, as different systems adapt to niches with different characteristics [13], and (iii) incommensurable: specialized to different niches and co-adapted to other systems, making cross-niche, cross-culture, or even cross-individual comparison at best challenging and at worst incoherent [14].

Rather than focus on the development of individual AI or psychological systems that optimize a single, general measure of intelligence (such as *g*, universal intelligence, or others that universalize a specific measure of individual task performance or cognitive capacity [15]), we argue on the basis of these observations that the analysis of intelligence should not focus on individual agents in isolation. Socially, an individual's intelligence should be analyzed in terms of the individual's role within its group, how the individual performs this role, and how the individual's performance in this role contributes to the group's success in the group's environmental niche. Ecologically, an individual's intelligence should be analyzed in terms of the interaction between the agent and its environment. Generally, intelligence is a property of a larger system of which the individual agent is a part.

## II.     Intelligence is a Disposition

In our initial definition, intelligence is defined as a certain sort of disposition. What is a disposition?[1] While we view dispositions as grounded in the causal powers of systems, we will not argue for this or any particular analysis of dispositions. But why think

---

[1] McKitrick (2003) gives some rules of thumb for determining whether a property is a disposition:
1. A disposition has a characteristic manifestation: a fragile glass breaks.
2. A disposition is triggered only under specific circumstances of manifestation: a fragile glass breaks when struck.
3. A disposition is associated with a counterfactual that is (typically) true of things with that disposition: "were the fragile glass dropped, it would break."
4. A disposition can be and often is referred to with dispositional locutions: fragility is "the disposition to break when struck."

We note that 'intelligence,' as the phrase is commonly used, bears all four of these marks. Using unanalyzed language (without unpacking the context from the agent): (i) an intelligent agent performs tasks successfully, (ii) an intelligent agent performs tasks successfully when the agent is motivated to perform those tasks successfully under conditions where the tasks can be performed successfully, (iii) if an intelligent agent were motivated to perform a task successfully under conditions where the task could be performed successfully, the agent would perform the task successfully, (iv) intelligence is the disposition to perform tasks successfully. Locutions of all four kinds are found throughout the literature on intelligence, in psychology, artificial intelligence, and philosophy. Our argument below implies a strengthening of (3).

intelligence is a disposition? We briefly motivate and defend understanding intelligence as a type of disposition, beginning with the idea that intelligence implies a type of potential. We analyze this kind of potential in terms of dispositions. Other ways of understanding this potential, in terms of capacities or mechanisms, either share the same conceptual grounds as our approach, converge on the same relevant properties of systems, or fail to explain such potential. This motivation and defense is not intended to be dispositive, as a full discussion of the dispositional status of intelligence would take the discussion too far afield. Rather, our argument assumes that intelligence is a disposition, and here we merely provide some initial motivation for thinking so, some of the problems associated with thinking of intelligence as a mechanism or as a capacity, and discuss how the view that intelligence is a disposition evades those issues.

Intelligence expresses the potential to act or think in some way as well as the acting or thinking itself. If a student is good at math, within the context of their environment and the constraints of what they've been taught, they have the potential to solve math problems. This implies not only that they in fact solve some math problem or other, or have solved such problems in the past, but that they could solve some math problems they haven't seen. Similarly, if an athlete is talented at tennis, again within the context of their environment and the constraints of their training, they have the potential to play tennis. This implies not only that they have in fact played before or are currently playing, but that they could play a variety of tennis games that they have never in fact played. In short, intelligence implies the potential to do certain things. As a consequence, while we define intelligence as the disposition to perform tasks, there must be some minimal threshold for task performance. A rock does not qualify as playing tennis extremely poorly, for example: a rock does not have the potential to do tennis-related activities at all. Without tennis potential of some sort, the rock cannot be considered to perform the task of playing tennis, no matter how many services are lobbed in its direction. The potential to do a certain minimum set of things is implied by intelligence. In order to analyze intelligence, some way to capture that potential is required.

To capture this potential of systems, we adopt a dispositional view. A disposition is one way to capture that potential. We say more about dispositions below, but as a first pass, a disposition implies that in particular contexts, a system that possesses the potential would act in certain ways (under a very broad conception of 'act'). Systems with dispositions can do things, that is, have the power to make something happen, whether that something is a movement, a thought, or some other effect.

The power to make something happen is often tied to causes and mechanisms. Why, then, isn't intelligence a mechanism? Here, we understand mechanisms as entities and activities causally organized such that they produce some phenomenon [16, 17]. Mechanisms do embody potential in the ways that they can produce outputs even if they are not currently doing so.

On the mechanist approach, intelligence is understood to be a mechanism. But mechanisms are identified by their entities, activities, and how those are causally organized. Hence, different causal organizations imply different mechanisms and hence different intelligences. But it is desirable to allow for variability in the causal organization of systems without thereby changing the intelligence of the system (or the judgment of intelligence of the system). For example, different people may have different brains– sometimes dramatically so–but they do not thereby differ in their intelligence. Or, a non-human animal and a human certainly have different brains, but may not differ in their intelligence regarding a particular task. Artificial neural networks certainly differ from humans and yet they, too, may be intelligent. Other cases of variation in mechanism but similarity or identity of intelligence can be readily imagined. In reply, it might be thought that intelligence is a mechanism in virtue of the mechanism satisfying some mechanism sketch [18]: there is a description of the mechanism that abstracts away from some but not all of the details about entities, activities, and causal organization, and it is in virtue of satisfying this description that some mechanism or other is intelligent. Since mechanism sketches leave out details that are used to type identify mechanisms, different types of mechanism can be intelligent. This reply fails on three counts. First, a mechanism sketch is not a mechanism and, depending on how much detail is left out, it is unclear if this proposal is different from our disposition view or the capacity view discussed next. Second, there may yet be different mechanism sketches that underlie different systems that exhibit similarly intelligent behavior, in which case the problem recurs. Finally third, and most relevantly for our view, the mechanism view assumes that the relevant variation across systems and the relevant sort of details left out to arrive at a mechanism sketch are details about the entities, activities, and causal organization of agents. But we maintain that intelligence is properly a property of agents-in-context. In reply, the mechanist can permit details outside the agent to be part of the system that is intelligent and permit abstractions that leave out some of those details. But then, there is an even wider amount of variation (because of all the diverse properties of contexts now permitted), and consequently the idea that even an abstract structure will apply to all systems seems implausible. Hence, we conclude that the mechanistic approach is ill-suited as an analysis of intelligence.

Why isn't intelligence a capacity? Capacities capture potentials as well, as the capacity to do X need not imply the doing of X. If we read capacity as analyzed in terms of dispositions or as implying dispositions in some way, then this is acceptable. If we see dispositions as analyzed in terms of or in some other way implying capacities, this is also acceptable, so long as this analysis or implication does not subvert our main thesis presented below, that dispositions imply true counterfactual conditionals.

However, if capacities are not analyzed in terms of nor imply dispositions, then the approach may miss out on some of the nature of intelligence. A capacity is an ability of some kind (to do or think X). Suppose we have an agent-in-context who poorly performs a task–say, an agent-in-context who can make pancakes but not tasty ones. Now, there is some minimal pancake-making potential required to attribute the disposition or the capacity to make pancakes to a system. This minimal threshold potential may not be the same for the capacity as it is for the disposition. Because capacities *sensu* non-dispositions are linked to the ability to do/think something, it is a success property. In the case of pancake making, this success property may include reference to the acceptable quality and flavor of the pancake. If that success property requires only some minimal threshold-for example, preparing any pancake, even if misshapen or barely edible-then the dispositional and non-dispositional capacity approaches may agree. We submit that our account *mutatis mutandis* would follow on such a capacity view. However, if the minimal threshold potential is high, such that e.g. making bad pancakes does not qualify, then the dispositional and capacity approaches diverge. But that is problematic, because intelligence is not a high-threshold success property; intelligence in the sense of the potential to perform a task covers the whole spectrum from some minimal threshold of dismal performance to whatever is optimal. A similar point applies if dispositions are understood as something other than capacities; however they are to be understood (such as in terms of powers), the proposed understanding must make room for this graded nature of intelligence.

Intelligence is best analyzed as a disposition. It is not well-analyzed as a mechanism because that understanding fails to identify intelligences despite differences in mechanism. It is not well-analyzed as a capacity (understood as something other than a disposition or some minimal threshold of performance at a task) because capacities do not admit of graded potentials. Intelligence implies the potential to do or think certain things, understood as a disposition of some system.

### III. Dispositions and Systems Identification

We have defined intelligence as a disposition to perform tasks. While we will not argue for a particular view of dispositions, we will argue that dispositions imply the truth of some subjunctive conditional (from here on, just 'conditional'), and use that implication in turn below to argue for (EI). Dispositions are dispositions of real systems in the world. As such, those systems will make some conditionals true, those related to when the capacities described by those dispositions are exercised. This simple insight will be expanded upon below and used to help defend (EI).

When does some whole composed of a set of parts acquire a disposition not possessed by the parts? We contend that the identification of the system that has a disposition relies on determining i) that the disposition is not possessed by any of the system's proper subsystems; ii) that the disposition is possessed by the system as a whole; and iii) that the system can be changed so that the disposition does not manifest (i.e., the system is susceptible to finks, discussed below) but does manifest under some conditions when the system is unchanged (i.e., the system is not an autofink, also discussed below).

Consider a system x consisting of a battery, a copper wire, and a bulb socket. To operate the system, we screw the bulb in, which depresses a catch on the bulb, exposing a conductive surface on the bulb and putting it in contact with a conductive surface on the interior of the socket. System x has a disposition to light bulbs, assuming that such a bulb is screwed into the socket. This implies that if a bulb were screwed into the socket, then the bulb would light.

Can this disposition be attributed to any of the constituent parts – the battery, wire, and socket – or subsystems – e.g., the battery and socket together – of the system? We argue not: the system with three parts is minimal as a possessor of the disposition in the sense that removing any of its parts prevents the manifestation of the disposition. To see this, consider each of the possible subsystems to which we might hope to attribute the disposition. Given the system consisting of battery, wire, and socket, there are 6 possible proper subsystems (3 subsystems with 1 part and 3 subsystems with 2 parts). Each such subsystem has dispositions, but careful canvassing of the cases reveals no subsystem with a disposition to light bulbs. The possible subsystems are:

1. The battery alone. The battery is disposed to produce a current through wires that are attached to it. But producing a current is not sufficient to light a bulb, and the disposition to produce current is not equivalent to possessing a disposition to light a bulb. The mere production of current will not light a bulb if, e.g., there is no socket to which the bulb attaches. Further, the removal of the wire from the system ensures that current will not be transmitted and the removal of the socket ensures that the bulb will not receive current.

2. The wire alone. The wire is disposed to conduct current between a source and a sink, such as between a battery and a socket, but this is insufficient to be disposed to light the bulb. In the absence of a source, current will not be conducted. In the absence of a current sink that connects the current sink to the transduction elements of the bulb, the wire cannot light the bulb.

3. The socket alone. The socket is disposed to receive current from a connected, live wire and transmit it to a bulb; but again, the possession of this disposition is not sufficient to be disposed to light the bulb. If the socket is not connected to a wire, then no current can be transmitted. If the wire is not connected to the battery, then the socket can transmit no current.

Similar analyses hold for the subsystems consisting of pairs of elements with a single one removed (i.e. the subsystems consisting of (4) the battery and the socket, (5) the battery and the wire, and (6) the wire and the socket). Consequently, the disposition cannot be attributed to any of the parts or their combination.

The individual components of the light-bulb system bring along their own dispositions which combine with those of the other components to produce the system's dispositions. The battery has the disposition to produce current; the wire has a disposition to conduct current; and the socket has a disposition to receive current and transmit current to a light bulb (and a similar analysis holds for the other strict subsystems of this system). None of these is equivalent to the disposition to light bulbs. Hence, the light bulb system satisfies (i). However, when each part is placed together and organized in the right way, the whole system is so disposed. So long as the battery holds a charge and is wired in series with the wire and socket, the light bulb system has a disposition to light bulbs that are screwed in. Hence the light bulb system also satisfies (ii).

The explanation of (iii) is more challenging. Intuitively, condition (iii) states that conditions exist under which the system will not light the bulb (for example, if the system is deliberately short-circuited), but the system will light the bulb if these conditions do not hold (in the absence of short circuiting and other kinds of interference, the bulb will light). Before unpacking this claim, we first defend the claim that a disposition entails a true conditional sentence that describes the conditions in which the system would respond. We begin with a statement of the simple conditional analysis from Lewis (1997)[19], that dispositions imply that if a system were to undergo a stimulus, it would respond a certain way. While problematic, we believe this thesis contains a kernel of insight. We revise the thesis into a simple conditional implication, that dispositions imply the truth of a conditional sentence. We then defend such an implication and turn back to a more formal explanation of (iii) after this defense.

## IV. Conditionals, Counterfactuals, and Counteractuals

To begin, consider the simple conditional analysis of dispositions. Lewis states the simple conditional analysis as "[s]omething x is disposed at time t to give response r to stimulus s iff, if x were to undergo stimulus s at time t, x would give response r" (Lewis 1997, p. 143). Here, we define the simple conditional analysis (SCA) as

(SCA) System S is disposed D to respond R to stimulus C iff if S were to undergo C, then S would respond R.[2]

How to understand the latter conditional in (SCA)? As Martin [20] establishes, it is neither necessary nor sufficient for some system S to have some disposition D that if S were to undergo C, then S would respond R. To simplify and following Martin, we turn our focus from the light bulb system to the disposition of a wire (S) to conduct electricity (D). If the wire (S) were connected to a charged battery (C), then the wire would conduct electricity (R). But now suppose that the wire is constructed such that when connected to a battery, the wire becomes dead, perhaps as the result of attaching a second wire to short circuit the first (Martin calls this an 'electrofink', or 'fink' for short). The wire has the disposition to conduct electricity, but when connected to a charged battery, it fails to do so. In other words, it is not necessary for the disposition that if the wire were connected to a charged battery, then it would conduct electricity.

Now consider the case where the wire does not possess the disposition to conduct electricity. Suppose that the wire is constructed such that when connected to a charged battery, the wire becomes live (Martin calls this a 'reverse fink'). The wire does not have the disposition to conduct electricity when connected to a charged battery but when connected to charged batteries, it does conduct electricity. This demonstrates that even though if the system S (the wire) were to undergo a stimulus C (be connected to a charged battery), then the S would respond R (conduct electricity), the system still does not possess the disposition—that is, the truth of the conditional is not sufficient for the disposition. In sum, the conditional in (SCA) is neither necessary nor sufficient for the disposition.[3]

A common reply to these counterexamples is to amend the (SCA) and provide further conditions. For example, an additional condition can be added to the (SCA) to arrive at the revised conditional analysis

---

[2] There are many problems with the simple conditional analysis of a disposition that we will not delve into here (see Lewis 1997; Choi and Fara 2021 for a review).

[3] Johnston 1992 and Bird 1998 also provide counterexamples of a different sort where some masking condition in the context of the system prevents the disposition's manifestation. The canonical example is of a fragile glass, which has the disposition to break when struck, that is packed in fluff. The fluff prevents the glass from breaking when struck. Nonetheless, the glass has the disposition to break when struck. There are also mimickers of dispositions (Lewis 1997) that look like dispositions but are not; the Hater of Styrofoam might destroy a styrofoam glass upon being struck even though the styrofoam glass has no such disposition.

> (RCA) System S is disposed D to respond R to stimulus C iff S has some property B and if S were to undergo C and has B for a sufficient period of time T, then S would respond R in T.[4]

This additional property B must be around for a sufficient period of time for the manifestation of the disposition. If the property is not around long enough, instantaneous finks or reverse finks would apply to the system.[5] However, these sorts of additional conditions generally fail. For, as Martin [21] has pointed out, there are innumerable ways that the conditional can fail, depending on the property B. In order to capture the full breadth of cases, B needs to be stated in sufficiently general terms. But the usual way to state B in such terms is by appeal to dispositions. In the case of the wire, the wire is disposed to conduct electricity when attached to a charged battery iff the wire has the property of being disposed to conduct electricity when attached to a charged battery and if the wire were attached to a charged battery and has the property of being disposed to conduct electricity when attached to a charged battery for some sufficient period of time, then the wire would conduct electricity in that period of time. But that is unsatisfactory as an analysis as it obviously contains the very disposition being analyzed.

For the defense of (EI), we maintain that a conditional of the form 'If S were to undergo C, then S would respond R' does follow from the disposition of a system. However, Martin's arguments challenge this claim. We will now argue that Martin's arguments are wrong because they imply a particular way of evaluating the truth of conditional sentences that we reject.

Dispositions may still imply the truth of a conditional even if they cannot be analyzed as such. (The shift to a disquotational thesis will be motivated below.) Consider a fragile glass. If a glass is fragile (read as: the glass has the disposition to break under certain stimuli), then "if the glass were struck by certain stimuli, the glass would break" is true. The truth of this conditional sentence may follow from the presence of the disposition, in which case the correct claim is a simple conditional *implication*:

---

[4] This is similar to Lewis' 1997 proposal.
[5] Maskers and mimickers (see footnote 3) may still be a problem, though we won't analyze whether this is so.

(SCI) If system S is disposed D to respond R under stimulus C then the sentence 'if S were to undergo C, then S would R' is true.[6]

Unlike (SCA), (SCI) does not state any analysis of a disposition. Rather, it states a disquotational thesis about a conditional sentence. Specifically, if some system has a disposition, then there is a conditional sentence that is true, namely the conditional describing the stimulus under which the disposition manifests. For example, for some glass that has the disposition to break when struck, the sentence 'if the glass were struck, then the glass would break' is true. But then, adopting a straightforward truth rule, since the sentence 'if the glass were struck, then the glass would break' is true, it is the case that if the glass were struck, then the glass would break.

However, considering the objections from Martin and others, the truth of the conditional sentence does not seem to follow from the presence of the disposition. Perhaps the glass is constructed such that if the glass were struck by that stimulus, the glass would not break because it hardens when struck by that stimulus (fink case).[7]

The reason for the shift to a disquotational implication is to rebut these considerations from Martin and others. Dispositions imply the truth of some conditional that describes contexts wherein the disposition manifests. That is, it is necessary that the conditional be true. This is particularly challenging for Martin and others who assert the existence of dispositions that can never manifest. To construct our reply, we will need to shift to a brief discussion of how to evaluate the truth of these conditional sentences.

How are the conditionals in (SCI) to be interpreted? An extensive literature exists on different interpretations of conditionals. In particular, there are dimensioned interpretations of the semantics of conditionals [22, 23, 24] that provide different accounts of the truth conditions for the very same conditional. To explain how the conditionals in (SCI) should be evaluated, we will compare counterfactual and counteractual readings of conditionals.

The evaluation of any conditional requires an assignment of individuals or kinds to the names that appear in the conditional (that is, the determination of the referents for the terms in the conditional). Consider a toy example. Let 'water' refer to H2O (as it is in the real world) and let the conditional be 'if water were to fall from the sky, then the stuff that falls from the sky would contain two hydrogen atoms'.

---

[6] This is an open sentence. What is true is the sentence where the variables are replaced by the relevant names.

[7] Perhaps the conditions are such that if the glass were struck in those conditions, the glass would not break because the Preserver of Glass would intervene and prevent the glass from breaking (masker case; see footnote 3).

A counterfactual reading of this conditional evaluates the conditional from the viewpoint of a world (scenario, etc.) considered as counterfactual. On a counterfactual reading, water cannot be other than H2O, because the denotation of 'water' is fixed by reference to the actual world. Then, the properties of that thing are evaluated in counter-to-the-facts worlds, those where water is falling from the sky even though it is not actually currently doing so. Since the denotation of 'water' is H2O and H2O contains two hydrogen atoms, the counterfactual reading of the conditional is true; if water considered counterfactually were to fall from the sky, then the stuff that falls from the sky would contain two hydrogen atoms.

Now take the counteractual reading. The counteractual reading evaluates the conditional from the viewpoint of a world considered as actual. On a counteractual reading, water can be something other than H2O; let's call it XYZ after Putnam [25]. The properties of the stuff denoted by 'water' are not fixed by the actual world's water. Suppose XYZ does not contain two hydrogen atoms. Then, the counteractual reading of the conditional is false; if water considered counteractually were to fall from the sky, then the stuff that falls from the sky would not contain two hydrogen atoms. But now suppose XYZ, whatever else it may be, does contain two hydrogen atoms. Then, the counteractual reading of the conditional is true; if water considered counteractually were to fall from the sky, then the stuff that falls from the sky would contain two hydrogen atoms. The referents of the terms in the counteractual reading are underdetermined and different interpretations can make the conditional true or false.

Now turn back to the glass case. In a counterfactual evaluation, 'glass' refers to the glass as it actually is. Because the counterfactual reading fixes the denotation of terms to the actual world and supposing glasses in the actual world do break when struck, the conditional is true. Real world glass (the sort that can be found in most kitchens) does not benefit from Preservers of Glass, who would intervene upon a striking to prevent a breaking; there are no supernatural interventions to save the glass; there are no mechanisms that, upon detecting a strike, suddenly harden the glass to make it unbreakable. On the counterfactual evaluation of the denotation of the terms in the conditional, the conditional is true because if the glass were struck, then the glass would break.

In contrast, in a counteractual evaluation, 'glass' refers to the glass as it is in some other world. There are numerous other such worlds. Take the case of glass made of unbreakabilium. If some thing is made of unbreakabilium, then that thing would not break when struck and, further, the thing does not have the disposition to break when struck. In that world, the conditional is false: it is not true that if the glass were struck, then it would break, because the glass is made of unbreakabilium. But it is also true that the glass does not have the disposition to break, because anything made of unbreakabilium is by hypothesis unbreakable, that is, does not have the disposition to break.

Now take a superunbreakabilium world. If some thing is made of superunbreakabilium, then that thing has the disposition to break when struck unless it is in fact struck, whereupon it does not break. Because the counteractual reading fixed the denotation of the terms to the world with superunbreakabilium, the conditional is false because if the glass were struck, then the glass would not break. Nonetheless, in that world, the glass purportedly has the disposition.

Which world is the relevant world under the counteractual determination of the denotations of terms in the conditional? The conditional is stated in (SCI). In (SCI), the system possesses the disposition. So the relevant evaluation of the conditional would be the superunbreakabilium counteractual: the glass does have the disposition to break when struck, but the conditional is false because if the glass were struck, then the glass would not break. To contravene (SCI), then, the objector must evaluate the conditional in (SCI) at a subset of counteractual worlds (for the example of the glass, these are counteractual worlds where the glass is made of superunbreakabilium and the like). The defender of the dispositions-imply-conditionals thesis, however, can reject counteractual interpretations of conditionals in the evaluation of (SCI).

How do we know the nature of the system such that we can be confident about the truth of a counterfactual conditional? The answer is to engage in inquiry, scientific or otherwise, about the system. We understand this as an exploration of the responses of a system when placed in different contexts. For example, when struck under a vast range of conditions (dropped from the countertop, hit with a hammer, etc.), glass will break. This set of contexts outlines the relevant counterfactuals within which the purported dispositions of the glass are to be evaluated. Granted that glass has the disposition to break when struck by a certain stimulus, there is some true conditional sentence implied by the disposition. The determination of which conditional is true is the conclusion of that inquiry. We conclude that the presence of a disposition implies the truth of a counterfactual conditional and not a counteractual conditional.

Consider again Martin's electrofink. The wire has the disposition to conduct electricity. Now, if an actual wire were to be connected to a charged battery, then the wire would conduct electricity. The electrofink is a phenomenon such that when a wire is connected to a charged battery, the wire is unable to conduct electricity. This makes the conditional false. But in order to evaluate the conditional as false, a counteractual reading is required: actual wires do not fink.

Our view implies that finked systems only have counteractual but not counterfactual dispositions. Let a finked system be a system with a fink: a system that possesses disposition D to respond R under stimulus C but were S to undergo C, S would not respond R. Finked systems imply that the conditional 'If S were to undergo C, then S would respond R' is false. As we have just argued, this conditional is true only under a counterfactual interpretation. Since the conditional is false, the conditional must be evaluated under a counteractual interpretation. Let a counteractual disposition be a disposition possessed only by systems that are different from how they actually are, and a counterfactual disposition be a disposition possessed only by systems given how they actually are. Since finked systems only make true a counteractual conditional and those counteractual conditionals require systems to be different from how they actually are, finked systems possess only counteractual dispositions.

Before moving on, one last objection. (SCI) maintains that dispositions imply the truth of a conditional sentence. As such, (SCI) is a disquotational thesis. Now, it might be thought that this disquotational thesis is far too strong: it seems that we are claiming that there exists some necessary conditional, that is, some conditional which is necessarily true. In other words, for a given disposition, there is some conditional which holds in every possible world. But regardless of the disposition, this seems to be false, for even on a counterfactual reading, there will be some world in which any given conditional thought to be implied by a disposition is false.

To reply to this objection, we distinguish de dicto from de re readings of our claim. The de dicto reading is that for any disposition, necessarily there exists some conditional or other that is true. The de re reading is that for any disposition, there exists some conditional which is necessarily true. The de re reading is much harder to satisfy because it implies a specific conditional is true. By way of illustration, compare:

> (i) If I live in a city, then necessarily, there exists someone who is my neighbor. (de dicto)
>
> (ii) If I live in a city, then there exists someone who is necessarily my neighbor. (de re)

(i) says that I have some neighbor or other if I live in a city, a fairly banal claim given the density of inhabitants in cities. In contrast, (ii) says that there is someone, some specific individual, who is my neighbor if I live in a city. That is a wildly strong claim. In the case of dispositions, the two claims are:

> (i*) If system S has disposition D, then necessarily there is some (C, R) such that if S undergoes C, then S responds R. (de dicto)
>
> (ii*) If system S has disposition D, then there is some (C, R) such that necessarily if S undergoes C, then S responds R. (de re)

The difference between the two claims is that in (i*), there is just some pair or other of stimulus and response such that if S undergoes C, then S responds R; whereas in (ii*), there is a particular pair of stimulus and response such that if S undergoes C, then S responds R. The latter claim is much stronger than the former, but we only intend the former claim.

In short, we are not claiming that for some disposition, there is always the same pair of stimulus and response such that the system responds in a certain way when undergoing that stimulus. We are claiming there will always be some pair or other such that the system responds under that stimulus. It need not be the same stimulus and response in every world, but in every world in which the system has the disposition, there will be some pair such that the system responds. This refinement of our claim avoids the objection that our standard for the presence of a disposition is too strong because it demands the same stimulus-response pair in every world in which the system has the disposition.

## V. Finks, Autofinks, and Systems Identification

Let us turn back to the analysis of grounds for dispositions. A finkish disposition is a disposition D to respond R (e.g., a disposition to light bulbs is a disposition to respond by producing light) when stimulated by C (e.g. the bulb is plugged into the socket), except when the system is stimulated by C, the system does not respond R. The failure to R-respond is the result of something, a fink, that prevents the manifestation of the disposition. A necessary property of a properly identified system is the possibility of another system that is not a subsystem of the first, but that in the presence of which, the first system's disposition is finked. In other words, the first system is a proper subsystem of the combination of the two systems and the combined system does not have the disposition. This property follows from the previous discussion: a system that has a counterfactual disposition can only fink counteractually. To get a system with a counterfactual disposition to fink, the system must change in some way–a part must be added (or subtracted), or the system must gain (or lose) a causal power. This changes the system from what it actually is.

This insight can aid in systems identification, i.e. the identification of a system that has some disposition. To identify a system that possesses a disposition is to identify the set of parts and their relations that together compose a system that possesses the disposition. For a given disposition, if the system is misidentified, then the system can seem to be an autofink. A fink is something that prevents the manifestation of the disposition of some system. An autofink is a system that is its own fink, that is, some system that has a disposition but the system itself prevents the manifestation of the disposition. In general, autofinks are impossible. Suppose some system S and some other system $S_0$. Suppose S has the disposition D such that if S were stimulated by some C then S would respond R. If $S_0$ is the fink of S, then there is some part p of $S_0$ (possibly $S_0$ itself) that prevents the manifestation of D. Suppose for contradiction that $S = S_0$. Since p is part of $S_0$ and $S = S_0$, p is part of S. But if p is part of S, then S can't have D, since it is not the case if S were stimulated by C then S would respond R. But ex hypothesi S has D. So $S \neq S_0$.

This proof requires the truth of some conditional that describes a situation in which the system manifests the disposition (viz., under some stimulus C, S manifests disposition D). As we have argued, this is a true conditional evaluated as a counterfactual, not a counteractual.

Autofinks can be used to diagnose whether a system has been correctly identified. For, given an arbitrary system S and disposition D, if S can be an autofink, then some supposition of the foregoing argument is false. In that argument, the supposition for contradiction was that $S = S_0$. But we can equally reject the hypothesis that S has D. Generally, whether to reject the system identification (i.e., $S = S_0$) or the presence of the disposition for an autofinked system will depend on the best explanation for the autofinkability of the system S. Call this the 'autofink test'. Autofinks are impossible, so a fink for a system which is just the system itself shows either that the system does not have the disposition or that some other system does have the disposition.

In objection, It might be argued that finks act in some way in order to prevent the response of the system, but autofinks don't act in failing to R-respond. In rebuttal, the claim is false. We can distinguish powerful from active finks. Active finks act to prevent the system's response. Powerful finks do not. A powerful fink does not need to act to prevent the system's response; rather, the existence of the fink is sufficient to disrupt the manifestation of the disposition. I may be disposed to stay awake after a cup of coffee, but in a hypoxic environment, I won't stay awake; the absence of oxygen, though, does not act to prevent the expression of the disposition because absences can't act.

When a system is properly identified, then the system is not an autofink. A finked system possesses the same disposition as the system but does not manifest the response when stimulated as the first system. (In fact, even modestly complex systems are susceptible to many finks—but always the result of a change to the system.) An autofink, in contrast, does not even possess the disposition. (And, consequently, the superunbreakabilium glass is an autofinked system.) Or, if we hold fixed the disposition, then the system must be improperly identified. The existence of a fink is an indicium of a properly identified system and the existence of an autofink of an improperly identified one. A disquotational way of putting this point is that finked systems make true only counteractual conditional sentences, whereas autofinked systems make true counterfactual conditional sentences.

Take the case of the light bulb. Suppose for each of the possible subsystems (battery alone, wire alone, etc.) we hypothesize that the subsystem has the disposition to light bulbs. Now suppose that the light bulb is screwed into the socket. For each of those subsystems, the subsystem will not light bulbs when the bulb is screwed in (this won't even be possible for the subsystems lacking the socket). If we consider each subsystem as the system and maintain that the system does have the disposition to light bulbs, then the system is an autofink. But autofinks are impossible, and so some supposition is false. Of course, the hypothesis that the subsystem has the disposition can be rejected. In the case of the light bulb, the best explanation of the seeming presence of the autofink is that the subsystem does not have the disposition.

## VI. The Argument from Grounds for Dispositions

Our argument for (EI) turns on an analysis of why some system possesses the disposition to perform tasks. We argue that the system possessing this disposition extends beyond the boundaries of the agent,[8] and ultimately the best explanation of the disposition is that the relevant system is an agent-in-context. This argument rests on the use of the foregoing autofink test.

When agents are removed from adaptive contexts, the ability to accomplish particular tasks drops precipitously. Conversely, when agents are placed in adaptive contexts, the ability to accomplish particular tasks increases. Hence, an agent's disposition to perform on particular tasks is context-dependent. Because different contexts result in different levels of performance for the same agent and the same context with different agents yields different levels of performance, the system possessing intelligent dispositions extends beyond the agent.

Consider three different cases of navigating subjects, adapted from Clark and Chalmers [26]. Otto has Alzheimer's Disease, which prevents him from navigating around

---

[8] We contrast external with internal and will use extended only in reference to the boundary of the agent. We avoid the dual of extrinsic and intrinsic, and we remain neutral on the relationship between the distinction between external and internal and the distinction between extrinsic and intrinsic.

Manhattan. In the first case, Otto must navigate from his hotel to the Met. Despite having Alzheimer's, Otto eventually makes it to the Met, albeit much more slowly than if he did not have the disease. Otto does not sit listlessly. He may make some wrong turns, be forced to backtrack, or take a longer route, but he ends up at the museum. Granted enough time and the grid structure of Manhattan streets, even Otto, who uses a simple strategy of randomly turning and avoiding getting run over or walking into the East River, will eventually get to the Met.

In the second case, Notto also has Alzheimer's but keeps a notebook to remember how to navigate around Manhattan. Notto is intelligent because he navigates when he is equipped with his notebook. Remove his notebook, however, and he navigates just like Otto without his notebook. Hence, the notebook is part of the system that navigates. Similar points apply to Notto's environment. Notto's intelligence is grounded both in his cognitive machinery (the representations and computations in his brain) and in his environment (the notebook, the city, and so forth). Properly speaking, both Otto and Notto are intelligent because both navigate. But Otto with the notebook is more intelligent than Notto without because Otto succeeds at navigation more often or with greater probability than Notto.

Finally, consider the third case of Wotto, also stricken with Alzheimer's and without notebook but who goes for a hike in the woods with his phone. While wandering in the woods, Wotto loses his phone. He wants to get back to his car but, having lost his phone, he can't navigate. In fact, he can only wander randomly because he is unable to use any environmental properties to navigate his way. Wotto is not intelligent (that is, cannot perform the navigation task; the issue of the scope of intelligence given some set of tasks is addressed below).

Compare Wotto with Otto. Neither have the external aids that Notto has. However, Otto can navigate while Wotto cannot. What is the difference? Wotto is on a hike in the woods, a relatively unstructured environment that fails to provide both cues for route-finding (and, hence, hints on how to get back to the car) as well as organization that can structure the search space to allow for some ability to navigate. Otto in contrast is wandering around Manhattan, which contains a regular street grid, vastly limiting the places that can be navigated and so permitting Otto the ability to navigate, albeit inefficiently. Obviously, to say that some system can perform some task involves some threshold or region of indeterminacy for the determination that the task can be performed. But given that threshold or region of indeterminacy–here, set to a level such that Otto is still navigating whereas Wotto is not–some systems are intelligent and others are not.[9]

The three cases of Otto, Notto, and Wotto illustrate various aspects of the context that must be included as part of the subject that is intelligent. Otto and Notto can both navigate as a result of different aspects of their contexts. Notto can navigate because of the tools that he has, namely the notebook, which buttress his memory and allow him to navigate to his goal efficiently. Otto does much worse, but he can still navigate–at a minimum, he can follow the contours of the regular street grid in systematic fashion until he arrives at his destination. Wotto, in contrast, can't rely on either his tools (because his phone is lost) or his environment (because he is stuck in the woods). As a consequence, Wotto can't navigate.[10]

Which system should we identify in each case as the system that possesses the disposition to navigate? Take Otto. Otto does not autofink: if Otto were to be placed in Manhattan, Otto would navigate. So far so good. Now consider Notto (i.e. Otto plus notebook). If Notto were to be placed in Manhattan, Notto would navigate. Hence, Notto is disposed to navigate. Notto is not an autofink. Double so far, so good. Finally consider Wotto. If Wotto were to be placed in the woods with neither notebook nor phone, Wotto would not navigate. This implies by the foregoing that either Wotto is not disposed to navigate or the system with the disposition to navigate has been misidentified.

The lessons that hold for Otto, Notto, and Wotto hold for intelligence generally. The disposition to perform a task will of course mean different things in different task contexts. Granted the disposition to perform a task, however, the purported system that has the disposition cannot be an autofink. If the system autofinks, then the system has been misidentified. But every agent alone is an autofink. Suppose some arbitrary agent has the disposition to perform some task. Suppose some other system, possibly the agent itself, is the fink of the agent. Suppose for contradiction that the agent is its own fink. The agent's disposition implies the truth of a counterfactual conditional; viz., if the agent were to undergo some stimulus, then the agent would perform the task. But generally agents can't perform tasks when they undergo some stimulus because various aspects of their context need to be present. Hence, agents autofink.

---

[9] Note that this does not make intelligence subjective. There could be objective reasons for setting the threshold (etc.) to some level to qualify as performing a task.

[10] Of course, if Wotto did not suffer from his cognitive problems, he could learn to use aspects of his woodsy environment to navigate (a scuffle here, a bent twig there; or perhaps the angle of the sun; and so forth).

Suppose that it is true that Otto and the notebook (or Otto, the regular street grid, and the notebook)–and possibly much else–are the relevant system that is intelligent. It does not follow from this that the proper bearer of the property of being intelligent is always an agent-in-context. In some contexts of use of 'is intelligent', such as Otto with his notebook, the proper subject is the agent-in-context, but in other contexts of use, the proper subject may still be the agent. Otto without the notebook, after all, can still navigate, just much more poorly, so why isn't Otto sans notebook the proper subject of intelligence in that context?

The spirit of the suggestion is agreeable to us. A context-relative treatment of the predicate 'is intelligent' may capture many of the cases that motivate our argument for extended intelligence. But there are some additional considerations that, we contend, warrant expanding the subject from just the agent to the agent-in-context in many cases. For example, Otto can navigate to some degree only because of other aspects of his environment–the regular street grid. In that case, the proper system is not Otto but rather Otto-in-Manhattan. Otto without Manhattan autofinks–Otto cannot navigate without the environmental structure provided by the regular street grid. The predicate 'is intelligent' is not (or not just) contextualized; in order to properly identify the system that is intelligent, the context must be included.

This lesson is amplified by considering how humans and other animals change their environment to buttress their abilities. Beavers can't beave in the absence of dams and Otto-without-notebook can't navigate (even poorly) in the absence of a regular street grid and the heuristics it enables. This sort of extra-organismic contribution to intelligence is trans-generational, inherited, and scaffolds the agent's contributions to the disposition to perform tasks [13,27]. Without them, no system is intelligent. These sorts of considerations of the role of parts of the environment in contributing to the ability to perform tasks motivates including those parts of the environment when carving out the part of the world that is intelligent.

However, from the extension beyond Otto of the parts of the system needed for Otto's disposition to perform tasks, we still cannot conclude that intelligence is also extended. Let the 'grounds' for a disposition be those parts of some system or other that allow the agent to perform tasks. The grounds for a disposition might be external to a system—that is, part of some other system (e.g., the environment)—without the implication that the system that possesses the disposition must be extended to include the grounds. A plant has the disposition to produce oxygen. This disposition requires carbon dioxide to manifest; but that does not imply that carbon dioxide is part of the plant. So, the objection goes, why should the fact that the grounds for intelligence extend beyond the agent imply that intelligence is a property of an agent-in-context system rather than an agent in isolation? Why think that the environment, society and other aspects of agents-in-contexts are not merely parts of the external grounds of the disposition of some agent to perform tasks as opposed to parts of the system that possesses the disposition? Call this the problem of external grounds for dispositions that are not parts of the system (the problem of external grounds for short).

This problem is profound. The autofink test can determine if a system has been misidentified. Repeated application of the test can generate a range of possible systems that have the disposition. Taking the minimal such system—the one with the fewest parts—will determine the identity of a system. However, this strategy won't separate the external grounds (the parts that contribute in some sense to the disposition but which are not part of the system) from the internal grounds (the parts that contribute to the disposition and are part of the system) of a disposition. Individual applications of the autofink test act as a lower bound, determining if some proposed system can possess a disposition. Repeated applications of the test will act as an upper bound, capturing all that is necessary for a system to manifest a disposition. But the latter will also include any external grounds for a disposition. We conclude the test is necessary but not sufficient for systems identification.

Is there another condition whose satisfaction in conjunction with the autofink test would be jointly sufficient for systems identification? We propose the following additional condition: for a given type of task and agent, the internal grounds of a disposition are those parts that i) are part of the minimal system that passes the autofink test and ii) are not present across types of tasks. The first condition is the iterated autofink test. The second condition, we propose, is an additional constraint that, in combination with the first condition, provides a set of conditions for resolving the internal and external grounds of dispositions. Call this the specificity condition.

The specificity condition excludes at least some external grounds of dispositions from inclusion as part of agents-in-contexts. Otto (with or without notebook) can't navigate anywhere let alone Manhattan without oxygen. But Otto also can't play tennis without oxygen, solve math problems without oxygen, or perform any of an enormous range of types of tasks. Consequently, while perhaps part of the grounds for the disposition to perform tasks, the oxygen needed for cellular functioning are external grounds, not internal grounds. As external grounds, oxygen is not part of the agent-in-context even though oxygen is needed to perform tasks.

The specificity condition crucially relies on both a type of task and an agent. An objection is that agents are present across types of task and, so, by the proposed additional condition, must also be left out as external grounds for dispositions! This objection though fails to reflect the presuppositions that are required to formulate the problem of external grounds. The motivation for conditioning on an agent is that the current concern regards internal vs. external grounds for some disposition of an agent-in-context. Which parts of the context are external grounds and which internal? To formulate that question, however, assumes some agent. So, even though the agent is present across types of task, the agent is not considered part of the external grounds for the disposition. For better or worse, we presuppose the agent. As we have argued, the grounds for the disposition to perform tasks reach beyond the bounds of the agent. The question now is how to draw the bounds of that extended system.

While the iterated autofink test is necessary for proper system identification, we do not claim the two conditions together are necessary or sufficient. There will instead be some range of cases, such as those of Otto, that adequately identify the intelligent system in virtue of satisfying both conditions. Other cases will pose troublesome and other conditions may prove valuable. Nonetheless, we contend that we have captured a range of central cases with our analysis.

What do we consider to be an agent? While a full reply to this issue is beyond the scope of this essay, we propose a minimal definition of an agent inspired by computational reinforcement learning theory. Minimally, an agent is a policy that maps states and rewards on to actions and the decision apparatus that implements the policy. This is a bare concept of an agent and we are open to expanding it. But whatever else might be considered part of the agent, our position maintains that parts of the context must be included in the bearer of the property of being intelligent.[11]

The thesis of (EI) maintains that intelligence is a property of agents-in-context. The autofink test indicates that intelligence is not a property of agents alone. The specificity condition specifies, for a given agent and type of task, what parts of the context are external grounds and not part of the agent-in-context and what are parts are internal grounds and part of the agent-in-context. There may be some vagueness or indeterminacy in drawing the boundary, a fact that does not undermine our thesis. The proper subject of intelligence remains an agent-in-context.

### VII. One Defense and Three Implications of Extended Intelligence

Having argued for the thesis of extended intelligence, we turn now to a defense and then three implications of the view. We defend the thesis of extended intelligence from the charge that granted the thesis, intelligence becomes maximally specific and so no longer useful. We also explore three implications of our view. First, intelligence is context-bound: since agents-in-contexts are the proper bearers of intelligent dispositions, intelligence itself cannot be divorced from the contexts of the agents whose behavior is estimated to be intelligent. Second, intelligence is task-particular. Different tasks imply different contexts and, since only agents-in-contexts are properly speaking intelligent, intelligence will be defined only with respect to some set of tasks or other. Finally third, since different agents will exist in different contexts and be posed with different tasks, intelligence is incommensurable: comparing across agents-in-contexts is ill-posed.

#### i. Intelligence as a Maximally Specific Predicate

---

[11] Another way to circumvent our view is to maintain that agents are simply identical to what we have been calling agents-in-context. We are sanguine about this metaphysical alternative.

On our view, intelligence is a property of agents-in-contexts. This may be thought to give rise to some problems. To begin, on our account, are there degrees of intelligence? Colloquially, some agents are said to be more intelligent than other agents. For example, Otto with the notebook is more intelligent than Otto without. Can our account make sense of claims like this?

In short, yes. On our analysis of intelligence, both Otto with the notebook and Otto without the notebook are disposed to navigate. Otto with notebook succeeds at navigating more often or with greater probability than Otto without. Even though Otto without notebook is worse at arriving at his destination, he can still navigate–that is, he still possesses some degree of intelligence. To make good on this claim, some measure of navigation performance is required to assess both Otto with and without notebook. Granted such a measure, such as the number of times within some period Otto with notebook arrives at his goal relative to Otto without, claims like 'Otto with the notebook is more intelligent than Otto without' can be evaluated. However, the references to Otto with the notebook or Otto without are elliptical; they refer to Otto with the notebook in Manhattan plus whatever else in the context needs to be included to assess Otto's intelligence. In addition, some measure of task performance is necessary to assess claims of relative intelligence or intelligence by degree. Some agent-in-context may be more intelligent than some other, relative to that measure.

However, this appeal to measures of task performance seems to run afoul of our analysis of intelligence. Any such measure, after all, will be part of the context within which the agent's performance is being assessed. So how, then, can there be a measure of task performance that doesn't further relativize the predicate 'is intelligent' or, more plainly, relativize intelligence? This problem ramifies on our proposed analysis. If intelligence attaches to agents-in-context, then any change in context, however miniscule, results in a new possessor of the property of being intelligent, including a change in the measure of task performance. Otto with a notebook in the morning is different from Otto with a notebook in the evening because of the change in context, and so their intelligences cannot be compared. Intelligence becomes a maximally specific property and 'is intelligent' becomes a maximally specific predicate.

Our reply is to note that not every change in the circumstances of an agent, as it were, is a change in context. An analogy to curve fitting in statistics will help clarify this reply. In curve fitting, there are variables (such as $x$ and $y$ in $y = m*x + b$) and there are parameters (such as $m$ and $b$ in $y = m*x + b$). A change in a variable is different from a change in the parameter. A curve is defined for a fixed set of parameter values over a varying set of values. To determine the goodness of fit of a curve to the data, the parameter values are chosen, held fixed, and evaluated over a range of values of the variables. To wit: the values of the slope $m$ and intersection $b$ are set to some value, and then different $y$ values are computed for different $x$ values. To determine the parameters of the best-fit curve, the overall goodness of fit of the curve is assessed for a range of values for that set of parameter values and then compared to the assessment of another curve defined by other parameter values.

We maintain a similar sort of distinction as that between variables and parameters holds for agents-in-contexts. Some properties of agents-in-contexts can change without thereby changing the agent-in-context. These are the variables. Other properties of agents-in-contexts cannot change without thereby changing the agent-in-context. These are the parameters. While we believe there are grounds for this distinction, we will not take a stand on which properties correspond to variables and which to parameters. But, holding fixed some such division into variables and parameters, intelligence is a property of agents-in-contexts. As a result, our view is consistent with both the view that most properties are variables to be included in the context as well as with the view that all are parameters (and, hence, every context is distinct from every other, no matter how small the difference).[12] Consequently, our view does not imply that intelligence is a maximally specific property or 'is intelligent' is a maximally specific predicate.

### ii.    Intelligence is Context-Bound

---

[12] Note that this is consistent with the view that some but not all properties are external grounds for dispositions. The current question regards what properties count toward distinguishing contexts; the other question regards what properties are properties of agents-in-contexts vs. properties external to agents-in-contexts.

The first implication of (EI) is that the context in which agents act is a necessary feature of intelligent systems. When denuded of the context to which an agent has adapted (usually by both inheritance and learning), an agent's ability to accomplish cognitive tasks will often drop precipitously. This is vividly illustrated by the historical failure of animal research to attribute cognitive sophistication to non-human animals, due to the animals' unsurprising failure to recapitulate tell-tale signs of cognition in experimental settings with wildly different cues or conditions than the animals' niche environments (de Waal 2016, p. 17–22). Conversely, when augmented with an environment specifically engineered to enhance an individual's idiosyncrasies, and ensuring the agent has the time, resources, and inclination to adapt to and control this environment, an agent's ability to accomplish tasks can be dramatically boosted. This is the context in which humans (inter alia) typically find themselves, and indeed accounting for the development of cognitive behavior and capacities in precisely such a context is a major outstanding problem for accounts of human evolution [13]. In short: an agent's disposition to perform on any task is not independent of context but rather depends on the context in which it is assessed; the ideal context for a given task depends on the agent in question.

Consider, side-by-side, a master Mousterian toolmaker (Fred) and a button pusher of the year 3000 (George). Fred is equipped with the most sophisticated tools of his time, and he has devoted years of practice and experimentation to their mastery. After many long years of apprenticeship (learning to precisely carve out and place stone cores, to precisely aim and strike with a secondary stone, and to choose the best chips and refine them by further chiseling into elegant and reliable axes), and years, longer still, of frustrating experimentation, Fred has developed a suite of novel techniques that allow him to manufacture the sharpest, most reliable axes available.

George, on the other hand, was raised in an automated city in the clouds, his every need cared for by attendant robots and his every question answered instantly by language models that distill the combined wisdom of human history. George is happy in his well-appointed, middle-class life, and he finds satisfaction in his work, which in practice consists of pushing a handful of buttons. These buttons control a sophisticated set of manufacturing protocols, and if pressed in the right sequence, they manufacture many tool parts at astonishing speed. The system has been designed to benefit from human input, but to be robust and operate safely even without it, so when George is inattentive the system is reduced in efficiency but it still churns out tool parts.

Who is more intelligent, George or Fred? We might be tempted to point at Fred. Through his life, he succeeded in one of the most challenging domains of his time (manufacturing hand axes) and consistently outperformed his peers in this domain, developing mastery and producing innovative new advances. But even in his domain of core competence (tool making), Fred's output is vastly inferior to George's, who can produce high quality tools of far greater consistency and utility than Fred's with much less effort.

The case for Fred becomes much more dire if we move from toolmaking to any modern test of general intelligence. Here, Fred is unlikely to fare even moderately well, as the cognitive skills he has fostered for success in his native environment—which aside from toolmaking may include skills such as foraging for food, bargaining and negotiating with Mousterian conspecifics, managing a household with his family group, but are likely to exclude formal training in analogical reasoning, reading comprehension, or formal mathematics—are unlikely to carry much weight on any modern aptitude test. George, on the other hand, spent six months at the culmination of his secondary education half-heartedly prepping for the space SATs with the help of his robot maid and uplifted dog. Of course, if we were to base our intelligence test on the skills necessary for success in Fred's world, we would find them indifferent to George, whose prosperity is predicated upon adaptation to an environment highly engineered for his success.

What if instead of comparing George and Fred head-to-head as adults, we raised them both as infants in George's world? Extrapolating forward to adulthood, we might expect Fred to be the more successful of the two, perhaps now running George's company or, at the very least, further innovating on the design of the tool parts George is only content enough to churn out. But this conclusion assumes the universality of Fred's potential as demonstrated by his success in the Mousterian culture. Fred's success in his own life occurs in virtue of the adaptiveness of his attributes to the context provided by his epoch. For example, in Fred's time, access to apprenticeships is limited (only certain members of certain families are permitted to apprentice) and predicated on an ability to submit unquestioningly to authority for many grueling years. After that period, reputational success as a master axe maker is predicated on an ability to balance impeccable craftsmanship with time spent developing and refining axes that outperform rivals', through trial and error. Thus Fred's success depends on details of his family history and status, his willingness and ability to suppress certain tendencies at certain phases of his life and express others at other phases, as well as the raw abilities resulting from the interaction between his biology, his experience, and his social context.

Any account of Fred's success will involve an analysis of this kind, and the details of his context as well as his abilities contribute to his differential success with respect to his conspecifics. Transplanted to George's world, some of these properties may still be adaptive—perhaps Fred's tolerance for failure and his knack for salesmanship may help— while others may not—an early predisposition to submission and a capacity for a high degree of motor control may be less well suited to certain post-industrial worlds. Depending on the nature of the traits and experience that Fred had, his success as a Mousterian axe maker might imply better or worse behavior on the tasks by which success is measured in George's post-industrial world.

Does the fact that agents alter their context, enhancing their own dispositions to perform at tasks, conflict with our analysis? Undoubtedly contexts are not stationary; context-shaping is an important part of evolutionary and cultural dynamics [12]. But changing the context changes both the nature of tasks and in many cases the nature of the agent itself. This is consistent with our account: an agent-in-context need not be static. Further, although an agent may alter its context, it always does so as an agent in some context. The facility with which an agent can change its context is predicated both on properties of the agent (how adapted is it to this context, so that it has access to means of changing it?) and properties of the environment (how supportive of the agent is the environment? how many affordances does it expose to the agent? how much potential for change does it have?).

EI implies that artificial general intelligence (AGI) in the strict sense is impossible. An objection to this implication contends that the development of a system that is better than the smartest human at all tasks that humans currently engage in (the typical unpacking of AGI in the sense of "weakly superhuman AGI") would disprove the claims made here. But this claim fails to account for the non-stationarity of the context that must occur to produce such an AGI. Such an AGI is predicated on the development of technologies, techniques, infrastructure, cultural development and so forth that do not exist in our present context C. Call the context in which they do exist C′. In context C′, human-mediated intelligence is now a property of humans-in-C′ rather than humans-in-C. This AGI and other systems like it are part of context C′ (like the tool-parts-producing system in George's world). The apparent force of this objection comes because it separates humans and the AGI from their contexts to force an uncharitable comparison.[13]

### iii.   Intelligence is Task-Particular

The second implication of (EI) is that intelligence is properly analyzed only for particular tasks. Can an agent perform a specific set of tasks and be considered intelligent, or must the agent be able to do many different things? On our view, agents-in-contexts are intelligent. But since a change in context can change the task, tasks are context-dependent. Take, for example, the task of completing a math exam with and without a calculator. These distinct contexts will engage different cognitive operations. The use of a calculator requires certain skills (entering digits, etc.) that are not required for completing the task without the calculator. But since tasks can change with context and intelligence is a property of agents-in-contexts, intelligence is task-particular.

In addition, the capacities of agents are shaped by the context in which they act. For example, bounded agents, those with limited memory, processing power, and so forth, will adapt to their local selective environments because they are resource constrained. Such local adaptedness suggests that intelligence is task-particular as different contexts will dictate different tradeoffs among constraints. So both considerations of contexts and agents imply that intelligence is task-particular.

The generality of the set of tasks on which the agent is disposed to perform well (given a context) appears in many discussions in AI. This is the sense in which "general" is used in AGI, which in the limit imagines agents that can perform well on any task. This sense of general is closely related to how "universal" is used, such as in Legg & Hutter's universal intelligence (Legg and Hutter, 2007b). Finally, it is reflected in Bostrum's "superintelligence", which is "an intellect that is much smarter than the best human brains in practically every field, including scientific creativity, general wisdom and social skills" (Bostrom 2006, p.11).

---

[13] A similar failure to account for the role of a non-stationary context for human intelligence may be at the root of the so-called 'AI effect', whereby problems that were considered hallmarks of human intelligence (deductive reasoning, strategic game playing, perceptual reasoning, language production, inductive and abductive scientific reasoning, etc.) come to be viewed in retrospect as essentially non-intelligent tasks ("mere computation") once they can be reliably be performed by AI systems (see McCorduck 2004, p. 423). Once the work required to successfully perform a task can be performed by the environment (in this case, a computer), the domain of 'human intelligence' is further restricted or refined. This shifts the goalposts on what tasks humans find challenging and interesting to perform.

How to differentiate the things at which an agent is good? Some agents develop general strategies that apply to many different tasks; others specialize. But this point requires a principled way to differentiate different tasks. This concern about how to count distinct cognitive capacities is general and remains underdiscussed and unresolved in the literature on cognition at large. One might adopt a formal principle for counting tasks: if the interpreted mathematical description of a task is distinct from and cannot be reduced to the interpreted mathematical description of another task, then the agential capacities on the two tasks are distinct.

Such a formal principle of task individuation is problematic. It is possible for different task descriptions to be available for the same task. One theorist might describe a task one way, and a different theorist another. There is also an underdetermination issue here, as two distinct task descriptions might coincide for the same range of observed or encountered contexts. However, this point only buttresses our position. Different descriptions of the same context can yield different assessments of how well some agent-in-context is performing. But this implies that intelligence, which is essentially partly assessed in terms of performance, must be task-particular. In fact, such mathematical descriptions will be insufficient to differentiate tasks, as many tasks may receive formally identical descriptions. To further distinguish tasks, details about the context must be included cf. [28]. So both the need to include contextual details and the inherent variability in how tasks are described imply the task particularity of intelligence.

If we mean that agents can vary greatly in intelligence based on their capacities in different environments, aren't we implicitly endorsing the proper domain of the applicability of the predicate 'is intelligent' to agents and not agents-in-contexts as our thesis (EI) claims?

By moving agents between contexts, a given agent can be part of different agents-in-context, which are disposed to successfully perform different kinds of tasks. With respect to a fixed set of reference tasks, agent A will perform better as a part of some agent-in-context systems than others. Suppose we have an agent A that, along with environments X and Y, can constitute an agent-in-context A-in-X or A-in-Y. A is adapted to X but not Y: in context X, A is disposed to perform a large set of tasks, but in context Y, A is disposed to perform a vanishingly small set of tasks. The possibility of A-in-Y means that A cannot be called intelligent simpliciter: in contexts other than X, A is not disposed to perform (nearly) any task. But A-in-X can be called intelligent, as this system is disposed to perform a large set of tasks without caveat. In this reading, calling agent A intelligent makes implicit reference to the context in which its successful, i.e. "A is intelligent" is shorthand for "A-in-X is intelligent."

### iv. Intelligence is Incommensurable

The third implication of (EI) is that the intelligence of agents (or of contexts) is incommensurable to other agents or contexts. By 'incommensurability' we mean that any measure of intelligence is relative to an agent and context. The disposition to perform tasks is incommensurable across agents because the agent and context jointly determine which behaviors are possible and so relevant to task performance. The adaptive argument for task-particularity also implies incommensurability because agents' behaviors are adapted to contexts and different agents may selectively adapt to their contexts. The implications of context-boundedness and task-particularity both imply that intelligence is incommensurable across agents.

This implication directly bears on attempts in psychology and artificial intelligence to define a single general measure of intelligence that can be used to compare agents. Consider for instance Spearman's *g* in psychology. This value is purportedly possessed to a greater or lesser degree by different people and to be revealed through the use of factor analysis on IQ tests [29]. But if the foregoing arguments are right, then there is no sense in which an agent has general intelligence to a greater or lesser degree: *g* merely measures performance in a set of contexts and on a set of tasks, with no necessary relationship with contexts or tasks in general. Instead, the tasks used to assess the system and the context in which the assessment is performed are essential to the assessment.[14]

Take tasks first. There is no 'general task' on which competence can be assessed. In which case, even if *g* is a real measure, the task assessed by *g* is not general in the relevant sense. In other words, saying of one agent that it has a greater degree of *g* than another is meaningless because there is no such thing as a disposition to perform tasks in general. *g* does not measure any general disposition; rather, it measures a specific disposition to perform on tests of intelligence in classroom-like settings.[15]

Further, the context in which a task is performed can undercut any measure of *g*. This is not only true in that some agents-in-contexts will be (say) more anxious about the test itself. As Gould (1996) notes (see e.g. p. 229-252), the particularities of the upbringing and cultural properties of agents can dramatically change the assessment. An agent placed in an unfamiliar context (such as being told to enter responses on a computer screen) will utterly fail at any purported 'intelligence' test in comparison to their performance in a familiar context (such as responding orally). These obvious points follow neatly from our view; an agent-in-context is intelligent in part because the context is an essential part of the relevant system that can be considered intelligent.

A specific consequence of this implication of our argument is that the notion of a truly general artificial general intelligence (AGI) system is incoherent unless the operative notion of generality is somehow qualified. This doesn't imply that current approaches to developing agents that can perform many tasks are fruitless and it doesn't mean that theoretical analyses like that of Legg and Hutter [9] are incoherent when analyzed properly. We argue that each of these cases should be analyzed as a particular task, with a particular context.

---

[14] Indeed, measures of intelligence like *g* that are derived from IQ or IQ-like tests have little predictive power even among subject populations for which they were originally designed on tasks and contexts beyond standardized testing in academic environments (Richardson and Norgate 2015). Their validity and predictive power for general task performance (and hence any basis for the claim that such tests measure intelligence) is noticeably weaker for other human populations (Belkhir 1994, Richardson 2002, Rosselli and Ardila 2003) and for AI systems (see Chollet 2019, section I.3.4).

[15] In addition, different agents-in-contexts may use different internal cognitive operations to perform tasks. But then, *g* is not assessing any internal operation either, because agents-in-contexts are doing fundamentally different things. A single score like *g* cannot assess intelligence in terms of tasks nor in terms of computations.

To see this, note that typical AGI task suites e.g. [4, 30, 31, 32, 33, 34, 35], as well as some AGI thought experiments, begin with an ansatz: that a specific measure captures or operationalizes what it is to exhibit general intelligence. This typically takes the form of an aggregate success rate on a set of tasks that are hand designed or selected from the literature to probe a specific range of "cognitive" abilities (in the case of suites) or expected performance with a weighting function motivated by simplicity or symmetry considerations (as in theoretical arguments like Legg and Hutter 2007b). After running a range of agents on the environment suite and measuring aggregate performance, the tentative conclusion is that agent A with the highest aggregate performance is the most intelligent.

We analyze this state of affairs as follows. While a particular task suite (or an infinite set equipped with a particular weight measure) is typically framed as a model of generality, it in fact defines a specific task – namely, maximizing performance on this specific evaluation. Let's assume that we can keep the context fixed across different agent evaluations, as would be required in standard rigorous experimental contexts. Then what the results of these (thought-)experiments show is task-particular evaluation. They do not run into the problem of incommensurability, because they are in fact restricted to a specific context (i.e. the specific training data, compute bounds, etc. used in a specific experimental configuration) and a specific task (i.e. the task defined by the task suite and often evaluated as e.g. the average performance over all of those tasks). What is measured in these cases is not general intelligence, but a variety of task-particular intelligence. If the goal of such an evaluation is to serve as a model of how the agent in question would perform in a human-like environment, the researcher's job is then to ensure that the specific context and task defined by the suite are as close as possible to human-relevant contexts and tasks.

A similar analysis holds for aptitude tests like IQ or SAT tests and other tests of human intelligence e.g. [36, 37]. These tests do not measure "general" intelligence. Rather, they measure performance on a specific task in a specific context, which can then be empirically – but never necessarily – connected to performance on other tasks in other contexts.[16]

---

[16] We are not alone in making this point. See e.g. Chollet 2019, Section II.1.2 for a discussion of generality claims in the context of tasks suite design. Indeed, many task suites (such as Osband et al. 2020) are explicitly designed to make such downstream development easier for practitioners.

The universalist view, on the other hand, is committed to commensurability. Say we have only two possible tests, tests I and II, and we weigh performance on both equally. According to a universalist view, A and B are equally intelligent if A can perform task I at success level 1.0 and II at 0 while B can perform task I at 0.5 and task II at 0.5. Or, consider a contrast between the Resilient Expert and Fragile Genius introduced by Crosby and Shevlin [38]. A Resilient Expert E is one that performs stably and reasonably well on a range of moderately difficult tasks, while a Fragile Genius G is brittle but performs excellently but only under a narrow set of conditions. There are many contexts C in which E produces good task performance, but very few contexts D in which G produces good task performance, but in these contexts task performance is excellent. A universalist might insist that we ask the question "is E or G more intelligent?"

To preserve the ability to make these kinds of statements in the face of the contextual nature of task performance, a universalist must qualify these statements in some way. This could perhaps be done by incorporating a precise notion of what is meant by context-invariant performance or by aggregating over all contexts in a neutral, theory-agnostic manner (both of which we believe are ultimately implausible) or by aggregating over contexts using a strategy that privileges some contexts over others.

We could put this in terms of generalized inequality, in which agents can be said to satisfy some criterion, but can't be linearly ordered, because the notion of performance is multidimensional (rather than a scalar). One can then compare performance by considering individual aspects of aggregate performance (A-in-X is better than B-in-Y at task I) or by imposing an aggregate metric (A-in-X is better than B-in-Y across all tasks using uniform weighting). But doing so requires the introduction of further machinery: a single task ranking does not follow that encapsulates all relevant comparison over all agents, contexts, and tasks.

Finally, one last implication of our view is that different agents-in-contexts doing different tasks can both be said to be intelligent. Contrary to what we claim, does this not imply that intelligence is commensurable? In reply, strictly speaking, saying that A-in-X-doing-Y is intelligent and B-in-P-doing-Q is intelligent are making different claims.

**VIII.   Conclusion**

Neither AI systems nor humans are intelligent or not. Instead, only agents-in-context are intelligent or not. Extended intelligence is not a linguistic thesis about the meaning of intelligence, an epistemological thesis about how we come to know how a system is intelligent, or a methodological thesis about how to assess intelligence. Rather, it is a metaphysical claim about the systems that can be called intelligent. The truthmakers of statements about the intelligence of agents are the states-of-affairs of those agents behaving in a context and so the truth of such claims is derivative from a more fundamental metaphysical truth about a system that extends beyond the mere agent. AI is intelligent because the artificial agent is situated in a context that includes a cultural milieu, possibly a human society, an environment – real or simulated – and this agent-cum-context is intelligent. Similarly, humans are intelligent because they are situated in a cultural milieu that contains other human and animal minds and this human-cum-context is intelligent. In short, intelligence is extended, a property of agents and their cultural, social, and environmental contexts.

What is the take-away from this view of intelligence? What does the view of extended intelligence buy us? We conclude with two evident impacts. First, we should not expect very much from intelligence tests, intelligence claims, or intelligence comparisons across agents. Removing the agent from the context leaves it entirely open-ended why some agent performs well or poorly on a task. To draw conclusions from such claims, more information is needed about the context within which the agent was acting. Second, to make progress in AI research, tuning of agents and of contexts is required. What makes a system smart is not just the amount of compute or the sophistication of the hardware. A system is also smart because of the context in which it decides and acts, and these contexts can be augmented, improved, and refined just as much as the algorithms or hardware.